# Improving Character Identity Consistency in Diffusion-Based Visual Story Generation through Identity Memory and Automatic Reranking

Maria-Emilia GRIGORE and Stelian SPÎNU

***Abstract*—Automatic visual story generation using diffusion-based generative models has become an increasingly important research direction in generative artificial intelligence. However, maintaining character identity consistency across consecutive frames remains an open challenge, as existing models often produce significant variations in facial traits, clothing, visual style, or anatomical proportions.**

**This paper proposes a modular pipeline for improving character identity consistency in multi-frame visual story generation using Stable Diffusion XL together with additional conditioning and automatic selection mechanisms. The proposed system integrates LoRA fine-tuning for character personalization, ControlNet for pose-based structural control, and IP-Adapter for visual identity transfer. Furthermore, a reranking mechanism based on Best-of-K candidate selection and persistent identity memory is introduced in order to automatically select the most visually consistent generated images.**

**The work also includes the integration and extension of the StoryDiffusion framework into a reproducible local workflow, together with additional improvements for generated frame selection. Performance evaluation is conducted using automatic metrics such as CLIP image similarity, DINO similarity and CLIP text-image alignment.**

**The experimental evaluation was conducted on a custom benchmark consisting of multiple characters and visual stories organized into different difficulty levels (easy, medium, and difficult). Both progressive variants of the proposed pipeline and extended versions of the StoryDiffusion framework were evaluated, including configurations such as SDXL baseline, Best-of-K, Identity Memory, and the Full Pipeline. The obtained results show that reranking, identity memory, and visual conditioning mechanisms help reduce identity drift across frames and improve the overall coherence and stability of the generated visual stories.**

## I. Introduction

Recent advances in diffusion-based generative models have significantly improved the quality of text-to-image synthesis. Models such as Stable Diffusion, DALL·E, and Midjourney are capable of producing visually appealing images from natural language descriptions and have accelerated research in automated visual content creation. Among the emerging applications of these models, visual story generation has attracted increasing attention due to its potential use in comics, animation, video games, educational materials, and interactive digital media.

Unlike single-image generation, visual storytelling requires the creation of a coherent sequence of images that depict the evolution of a narrative while preserving the visual identity of the characters across multiple frames. Although modern diffusion models can generate highly detailed images, maintaining a consistent character identity throughout a story remains a challenging problem. Character appearance may gradually change between frames, resulting in variations in facial features, clothing, proportions, colors, or overall artistic style. These inconsistencies become more pronounced in longer stories and in scenes involving significant changes in pose, viewpoint, illumination, or composition.

The limitations of current approaches are largely related to the fact that most diffusion models are designed to generate individual images independently rather than maintaining a persistent representation of a character throughout a sequence. Even when additional conditioning mechanisms such as LoRA, IP-Adapter, or ControlNet are employed, identity drift may still occur, especially in complex scenarios. Furthermore, selecting the most suitable image among multiple generated candidates often requires manual inspection, making the process time-consuming and difficult to reproduce.

Another challenge is the lack of standardized benchmarks specifically designed for evaluating character consistency in multi-frame visual stories. Existing studies frequently focus on human identity preservation or evaluate only a limited number of examples, making objective comparisons between methods difficult.

Motivated by these limitations, this paper investigates the problem of character identity consistency in diffusion-based visual storytelling, with a particular focus on cartoon-style characters. A modular generation pipeline is proposed, combining SDXL, LoRA, IP-Adapter, and ControlNet with two additional mechanisms: an automatic Best-of-K

M.-E. GRIGORE is a Master's student in Artificial Intelligence for Defense and National Security at the Faculty of Information Systems and Cyber Security, Military Technical Academy "Ferdinand I", Bucharest, Romania (e-mail: maria.grigore@mta.ro).

S. SPÎNU is with the Faculty of Information Systems and Cyber Security, Military Technical Academy "Ferdinand I", Bucharest, Romania (e-mail: stelian.spinu@mta.ro).

reranking strategy and a persistent Identity Memory module. In addition, the StoryDiffusion framework is extended and adapted for reproducible benchmarking. To evaluate the proposed methods, a dedicated benchmark containing multiple characters and stories organized into different difficulty levels was created, enabling a systematic analysis of character consistency across a wide range of narrative scenarios.

The main contributions of this paper are summarized as follows.

1. An experimental study was conducted on the combined use of SDXL, LoRA, IP-Adapter, and ControlNet for multi-frame visual story generation, highlighting the limitations of purely frame-wise identity conditioning and the occurrence of identity drift in longer sequences.
2. A modular framework was designed and implemented for multi-frame story generation, integrating explicit character representation, identity conditioning, automatic frame selection, and evaluation components, enabling systematic comparison of different consistency-preservation strategies.
3. An automatic reranking approach was introduced, generating multiple candidate images for each frame and selecting the optimal result based on a combination of identity, attribute, and text-image alignment scores.
4. A persistent identity memory mechanism was proposed to propagate character-specific visual information across frames, reducing identity drift and improving character consistency throughout the generated stories.
5. A custom benchmark composed of multiple characters and stories organized into easy, medium, and difficult difficulty levels was created for the objective evaluation of character consistency in multi-frame visual storytelling.
6. The original StoryDiffusion framework was adapted for local execution, automated experimentation, and reproducible benchmarking, enabling a direct comparison between the proposed methods and existing approaches.

The implementation discussed in this paper is publicly available on GitHub at the following repository: https://github.com/MerryWex/character-consistency-visual-storytelling.

## II. Related Work

### A. Diffusion Models for Visual Story Generation

Diffusion-based generative models have become the dominant approach for text-to-image synthesis due to their ability to generate high-quality and semantically coherent images from natural language descriptions. Among the most influential open-source solutions, Stable Diffusion introduced a latent diffusion architecture capable of producing detailed images while remaining computationally efficient [1].

Subsequent developments led to the introduction of Stable Diffusion XL (SDXL), which improved semantic understanding, image quality, and visual realism through larger architectures and dual text encoders [2]. Although SDXL significantly improves image generation quality, it still treats each image as an independent generation process and does not provide an explicit mechanism for maintaining character identity across multiple frames.

To address this limitation, StoryDiffusion was proposed as a framework specifically designed for visual storytelling and multi-frame generation [3]. Unlike conventional text-to-image systems, StoryDiffusion introduces mechanisms for sharing visual information between consecutive frames through Consistent Self-Attention. This allows identity-related features to be propagated throughout a sequence, improving character continuity. However, identity drift may still occur in longer stories or in scenes involving substantial changes in pose, viewpoint, illumination, or composition.

### B. Identity Preservation Methods

Recent research has proposed several approaches for preserving visual identity in diffusion models. Some methods rely on fine-tuning techniques, while others transfer identity information directly from reference images.

InstantID employs facial embeddings extracted from reference images to guide the generation process without requiring extensive fine-tuning [4]. The method is capable of producing identity-preserving results from only a small number of examples, making it suitable for rapid personalization tasks. However, its performance is primarily optimized for realistic human faces.

PhotoMaker extends this idea by combining multiple reference images of the same subject to construct a more robust identity representation [5]. By aggregating information from different viewpoints and expressions, the method improves identity preservation and reduces sensitivity to individual reference images.

In parallel, techniques such as LoRA, IP-Adapter, and ControlNet have become widely used for personalization and controllable image generation [6-8]. LoRA enables efficient character-specific fine-tuning, IP-Adapter introduces image-based conditioning, while ControlNet provides structural guidance through auxiliary inputs such as poses or edge maps. These methods have demonstrated strong performance for single-image generation and identity transfer tasks, but maintaining stable character representations across entire visual stories remains challenging.

### C. Research Gap

Despite the significant progress achieved by recent diffusion-based approaches, character consistency in multi-frame visual storytelling remains an open research problem. Most existing methods focus on realistic human identities, short image sequences, or single-image personalization tasks. Their performance often degrades when stories become longer or when characters are depicted in diverse poses, environments, and narrative situations.

Furthermore, many approaches depend heavily on prompt engineering, manual image selection, or expensive fine-tuning procedures. Existing systems also lack explicit mechanisms for continuously updating and reusing identity information throughout a generated story. In addition, there is currently no widely adopted benchmark specifically designed for evaluating character consistency in cartoon-style visual storytelling.

These limitations motivate the approach proposed in this paper, which combines identity conditioning, automatic candidate selection, and a persistent Identity Memory mechanism within a modular visual storytelling pipeline. The objective is to improve character consistency while maintaining full automation and reproducibility of the generation process.

## III. Methods

### A. Modular Visual Storytelling Pipeline

The primary contribution of this work is a modular framework designed to improve character consistency in multi-frame visual storytelling. The proposed system combines diffusion-based image generation with identity conditioning, automatic candidate selection, and persistent character representations. The objective is to preserve the visual identity of cartoon-style characters while allowing variations in scene composition, pose, viewpoint, and narrative context.

Unlike conventional text-to-image approaches, which generate each image independently, the proposed pipeline treats character identity as information that should persist throughout the entire story. Consequently, the generation process is formulated as a sequence-level task rather than a collection of unrelated image synthesis operations.

The framework is organized as a set of independent but interconnected modules responsible for character representation, prompt construction, image generation, candidate evaluation, and identity propagation. This modular design allows individual components to be evaluated separately and facilitates systematic experimentation with different identity-preservation strategies.

Each character is represented through a dedicated profile containing identity prompts, reference images, visual attributes, and consistency constraints. These profiles separate stable character characteristics from scene-specific narrative descriptions. As a result, modifications in background, action, or camera perspective can be introduced without requiring the entire character description to be rewritten for each frame.

Story descriptions are stored using structured JSON files. Each story contains a sequence of narrative frames together with generation parameters and identity-related information. A Prompt Builder module automatically combines character descriptions with scene-level instructions, ensuring a consistent textual structure throughout the generated sequence.

Image generation is performed using SDXL [2] as the base diffusion model. To improve character consistency, three complementary conditioning mechanisms are employed. Character-specific LoRA adapters [6] provide visual specialization learned from reference images, IP-Adapter [7] introduces image-based identity conditioning during inference, and ControlNet [8] supplies structural guidance through pose information. These components address different aspects of the generation process and therefore operate in a complementary manner.

LoRA is primarily responsible for learning character-specific appearance patterns from a relatively small training dataset. By modifying only a low-rank subset of model parameters, LoRA enables efficient character personalization without requiring full model retraining. IP-Adapter introduces visual information directly from canonical reference images, allowing the generator to rely not only on textual descriptions but also on explicit visual examples. ControlNet is used to constrain body posture and scene structure, reducing undesired variations in character positioning across frames.

Although the combination of SDXL, LoRA, IP-Adapter, and ControlNet significantly improves consistency, preliminary experiments revealed that identity drift still occurs in longer stories. Character appearance may gradually change between frames, particularly when scenes involve substantial changes in viewpoint, lighting conditions, or interactions with other objects. To address this issue, the proposed framework introduces a persistent Identity Memory mechanism.

Identity Memory is initialized from one or more canonical reference images associated with a character. Visual embeddings are extracted using a CLIP-based Appearance Encoder [9], and the resulting embeddings are averaged and normalized to obtain a single identity representation. This representation serves as a persistent reference used during candidate evaluation.

For each narrative frame, the system generates multiple candidate images instead of relying on a single stochastic generation. This strategy follows a Best-of-K paradigm, where several alternative solutions are produced using identical prompts but different sampling trajectories. The resulting candidates are subsequently evaluated by an automatic reranking module.

Candidate evaluation combines three complementary criteria. The identity score measures similarity between the candidate image and the stored identity memory using cosine similarity in the CLIP embedding space. The prompt score estimates the semantic alignment between the generated image and the textual description of the scene. The attribute score verifies whether predefined visual characteristics, such as dominant colors, clothing elements, or accessories, are preserved.

The final ranking score is computed as

$$Score_{final} = 0.5 \cdot Score_{identity} + 0.1 \cdot Score_{prompt} + 0.4 \cdot Score_{attributes}$$

The weighting scheme was determined empirically through preliminary experiments. Identity preservation receives the highest weight because maintaining a stable character representation constitutes the primary objective of the framework. Attribute consistency also plays an important role, as many observed failures involve changes in clothing, colors, or accessories. Prompt alignment receives a smaller weight but remains necessary to prevent the selection of visually consistent images that do not correctly represent the intended scene.

After the highest-scoring candidate has been selected, its embedding is incorporated into the identity memory. The memory update is performed using a momentum-based formulation:

$$M_{t+1} = \alpha M_t + (1 - \alpha) E_t \,,$$

where $M_t$ denotes the current memory representation, $E_t$ is the embedding extracted from the selected image, and α controls memory persistence. In the proposed implementation, $\alpha = 0.85$.

The resulting embedding is subsequently normalized:

$$M_{t+1} \leftarrow \frac{M_{t+1}}{\|M_{t+1}\|_2}.$$

This mechanism allows the memory to evolve gradually throughout the story while preserving the dominant visual characteristics of the character. Consequently, the system becomes more robust to pose changes, viewpoint variations, and complex narrative transitions. The overall architecture of the proposed modular framework is illustrated in Fig. 1.

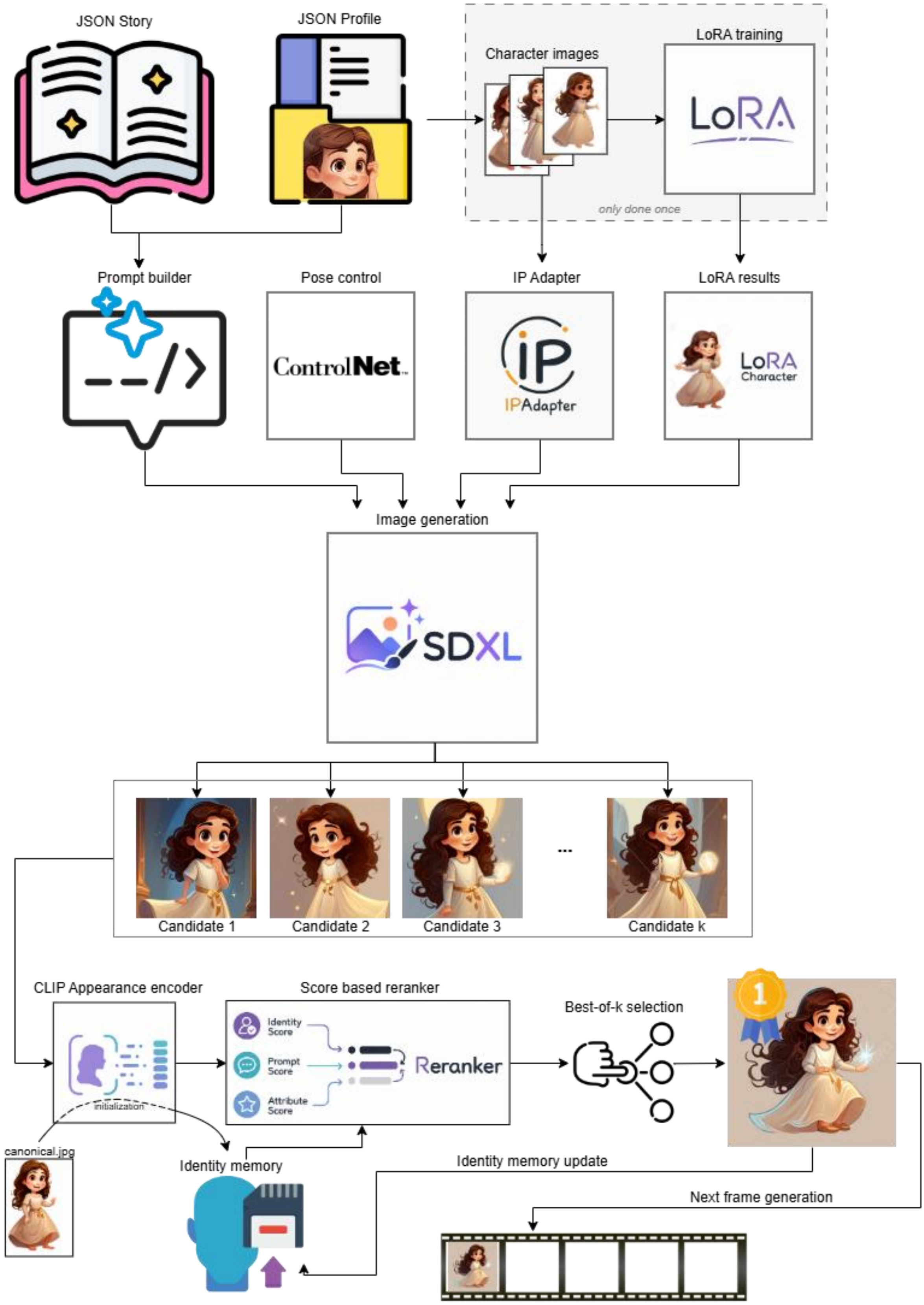

Figure 1. Overview of the proposed modular visual storytelling pipeline

## B. StoryDiffusion Extension

The second contribution of this work consists of extending the StoryDiffusion framework [3] and adapting it for reproducible experimentation and benchmarking. While StoryDiffusion already improves long-range consistency through Consistent Self-Attention, preliminary experiments revealed several limitations when generating longer stories involving cartoon-style characters.

The most significant limitation was the gradual degradation of character identity over time. Although self-attention propagation helps preserve certain visual characteristics, the framework does not maintain an explicit persistent identity representation. As a consequence, character appearance may drift progressively as the narrative evolves.

Another limitation concerns the stochastic nature of diffusion models. The original implementation generates a single image for each frame, meaning that the final result depends heavily on a particular sampling trajectory. In practice, different generations produced noticeably different levels of character consistency despite using identical prompts.

Furthermore, the original framework does not provide an automated mechanism for evaluating identity preservation or selecting the most suitable frame among multiple alternatives. As a result, qualitative inspection remains the primary evaluation method, making large-scale comparisons difficult.

To overcome these limitations, the StoryDiffusion framework was restructured into a fully configurable command-line workflow supporting batch execution, automated benchmarking, and reproducible experimentation. This modification enabled large-scale evaluation across multiple characters, stories, and difficulty levels.

The analysis performed using the proposed pipeline identified three mechanisms that produced the largest improvements in consistency: Best-of-K generation, automatic reranking, and Identity Memory. These components were therefore integrated into StoryDiffusion to evaluate whether their benefits generalize beyond the custom pipeline.

The first extension introduces Best-of-K candidate generation. Instead of accepting a single generated image, the framework produces multiple alternatives for each narrative frame. This increases the probability of obtaining at least one candidate that simultaneously preserves identity and satisfies narrative constraints.

The second extension incorporates the reranking strategy previously described. Candidate images are evaluated using the same identity, prompt, and attribute scores employed in the proposed pipeline. This allows StoryDiffusion to automatically select the most suitable frame without manual intervention.

The third extension introduces Identity Memory into the StoryDiffusion workflow. While the original framework relies exclusively on internal attention propagation, the extended version maintains an explicit CLIP-based identity representation that is updated throughout the generated sequence. This additional memory provides a stable identity reference even when substantial scene variations occur.

Finally, IP-Adapter conditioning [7] was integrated into StoryDiffusion to provide explicit visual identity references during inference. The resulting framework combines self-attention consistency, image-based conditioning, automatic candidate selection, and persistent identity memory within a unified generation process. A comparison between the original StoryDiffusion workflow and the proposed extended version is presented in Fig. 2.

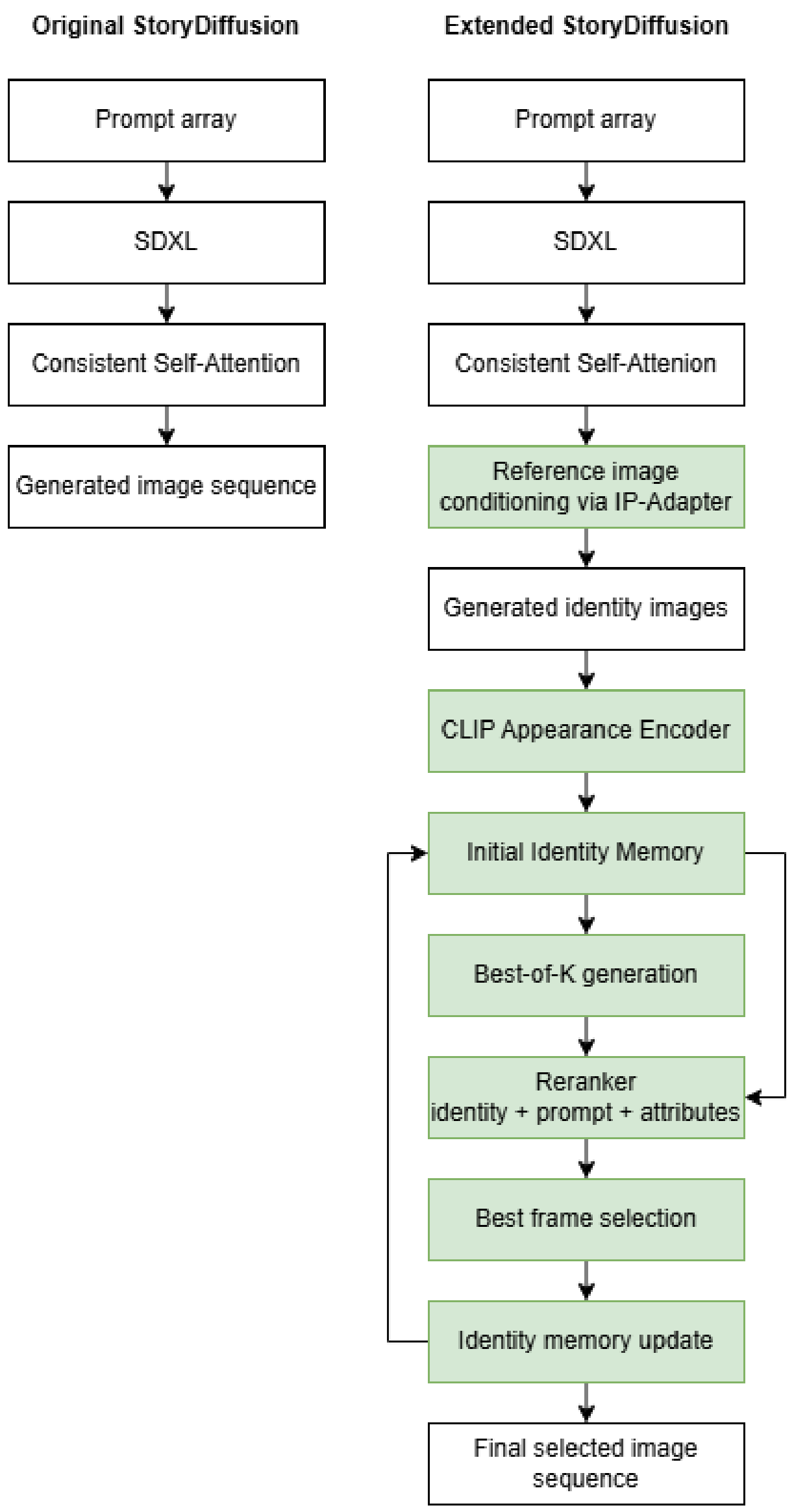

Figure 2. Comparison between the original StoryDiffusion workflow and the proposed extended version

## IV. RESULTS

### A. Experimental Setup

The experimental evaluation was designed to assess the effectiveness of the proposed mechanisms for preserving character identity in multi-frame visual storytelling. All experiments were conducted using an NVIDIA A100 80GB GPU and the PyTorch framework. SDXL was employed as the base diffusion model for both the proposed pipeline and the StoryDiffusion extension.

To ensure a fair comparison, all methods were evaluated using the same characters, stories, generation parameters, and inference settings. Controlled random seeds were used throughout the experiments to minimize stochastic variations and isolate the impact of architectural differences.

Since no standardized benchmark currently exists for evaluating character consistency in cartoon-style visual stories, a dedicated benchmark was created for this work. The benchmark contains six cartoon-style characters and multiple stories organized into three difficulty levels: easy, medium, and difficult. The six characters included in the experimental benchmark are presented in Fig. 3.

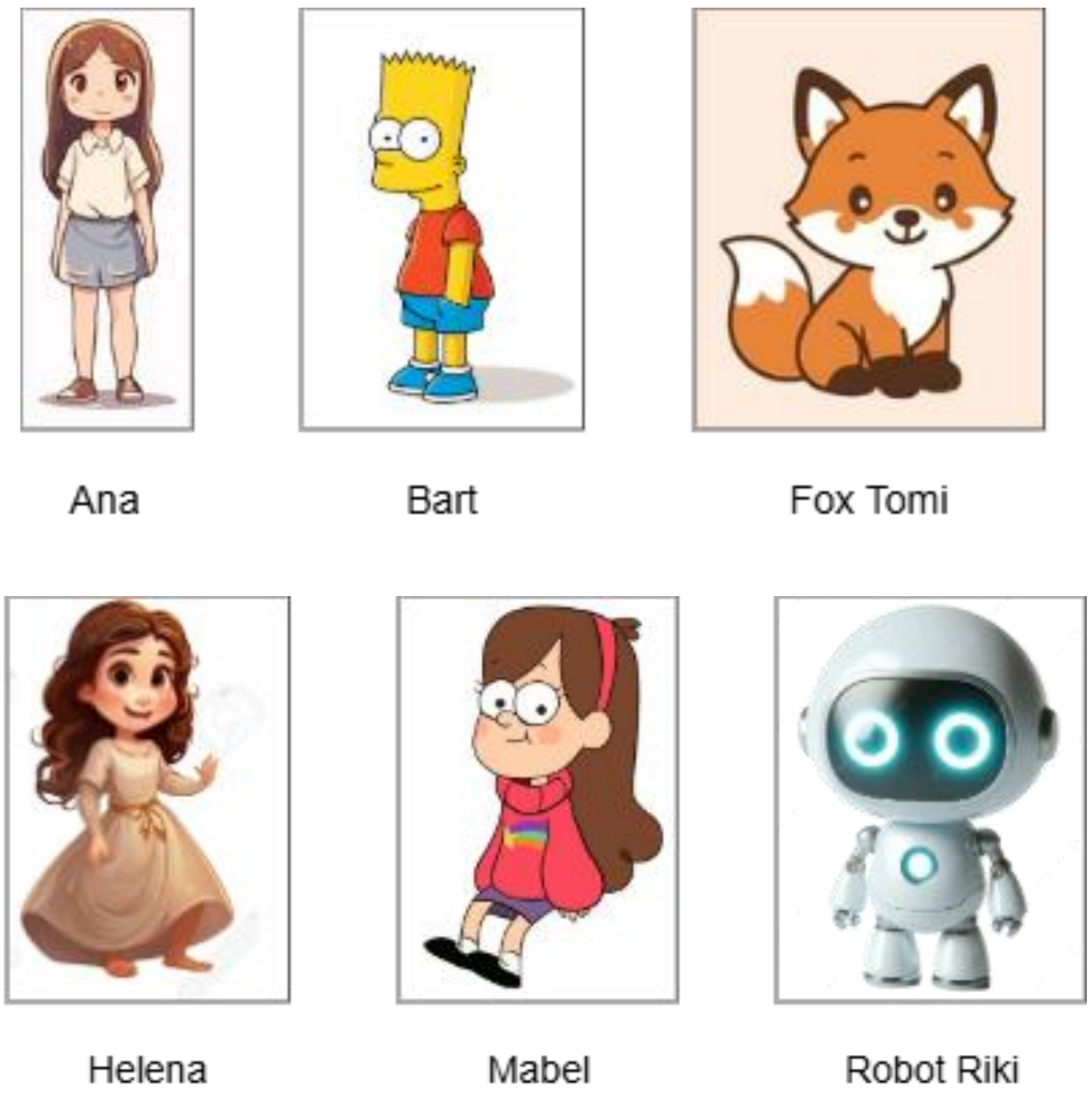

Figure 3. Characters included in the experimental benchmark

The stories introduce progressively larger variations in pose, facial expression, viewpoint, scene composition, and environmental complexity.

Three families of methods were evaluated:

– Original StoryDiffusion (text-only baseline);
– Extended StoryDiffusion;
– Proposed pipeline.

More than 160 experimental runs were performed, including both ablation studies and direct comparisons between methods.

### B. Evaluation Metrics

Two categories of metrics were used during evaluation. Internal metrics were employed by the reranking mechanism, while external metrics were used exclusively for benchmarking and objective comparison.

The internal evaluation combines three complementary components: identity preservation, prompt alignment, and attribute consistency. These scores are used during Best-of-K candidate selection.

For external evaluation, three widely used metrics were adopted:

– CLIP Image Similarity (CLIP-I);
– DINO Similarity;
– CLIP Text Alignment (CLIP-T).

CLIP-I and DINO measure visual and structural consistency with respect to the reference character, while CLIP-T evaluates the semantic alignment between generated images and textual descriptions.

### C. Quantitative Results

Table I summarizes the overall performance of all evaluated methods.

The quantitative results demonstrate that identity-conditioning mechanisms substantially improve character consistency compared to text-only generation approaches. The original StoryDiffusion baseline obtains the lowest overall score, confirming the limitations of relying exclusively on textual descriptions for multi-frame generation.

Introducing IP-Adapter, Best-of-K selection, and Identity Memory consistently improves performance across all evaluation metrics. The largest improvements are observed for identity-related measures, indicating that explicit visual conditioning and persistent character representations effectively reduce identity drift.

TABLE I. OVERALL BENCHMARK RESULTS

| Family | Method | CLIP-I | DINO-I | CLIP-T | Mean Score |
|---|---|---|---|---|---|
| StoryDiffusion original | Textual prompt only | 0.674 | 0.535 | 0.278 | 0.496 |
| StoryDiffusion extended | Baseline | 0.805 | 0.587 | 0.286 | 0.559 |
| | Best-of-K | 0.805 | 0.585 | **0.297** | 0.562 |
| | Identity Memory | 0.836 | 0.613 | 0.289 | 0.579 |
| | Full | 0.827 | 0.607 | 0.287 | 0.574 |
| Proposed pipeline | SDXL only | 0.706 | 0.461 | 0.269 | 0.479 |
| | Best-of-K | 0.754 | 0.5 | 0.278 | 0.511 |
| | IP + ControlNet | 0.782 | 0.551 | 0.283 | 0.539 |
| | Full pipeline | **0.848** | **0.629** | 0.287 | **0.588** |

The proposed pipeline achieves the best overall performance, obtaining the highest CLIP-I, DINO-I, and aggregate benchmark scores. The complete configuration reaches an overall score of 0.588, outperforming both the original StoryDiffusion implementation and the extended variants. An overall comparison of the evaluated methods based on the aggregate benchmark score is shown in Fig. 4.

The results also suggest that identity preservation and semantic alignment are not perfectly correlated. Several methods achieve strong identity scores while producing only marginal improvements in CLIP-T, highlighting the trade-off between preserving character appearance and maintaining scene flexibility.

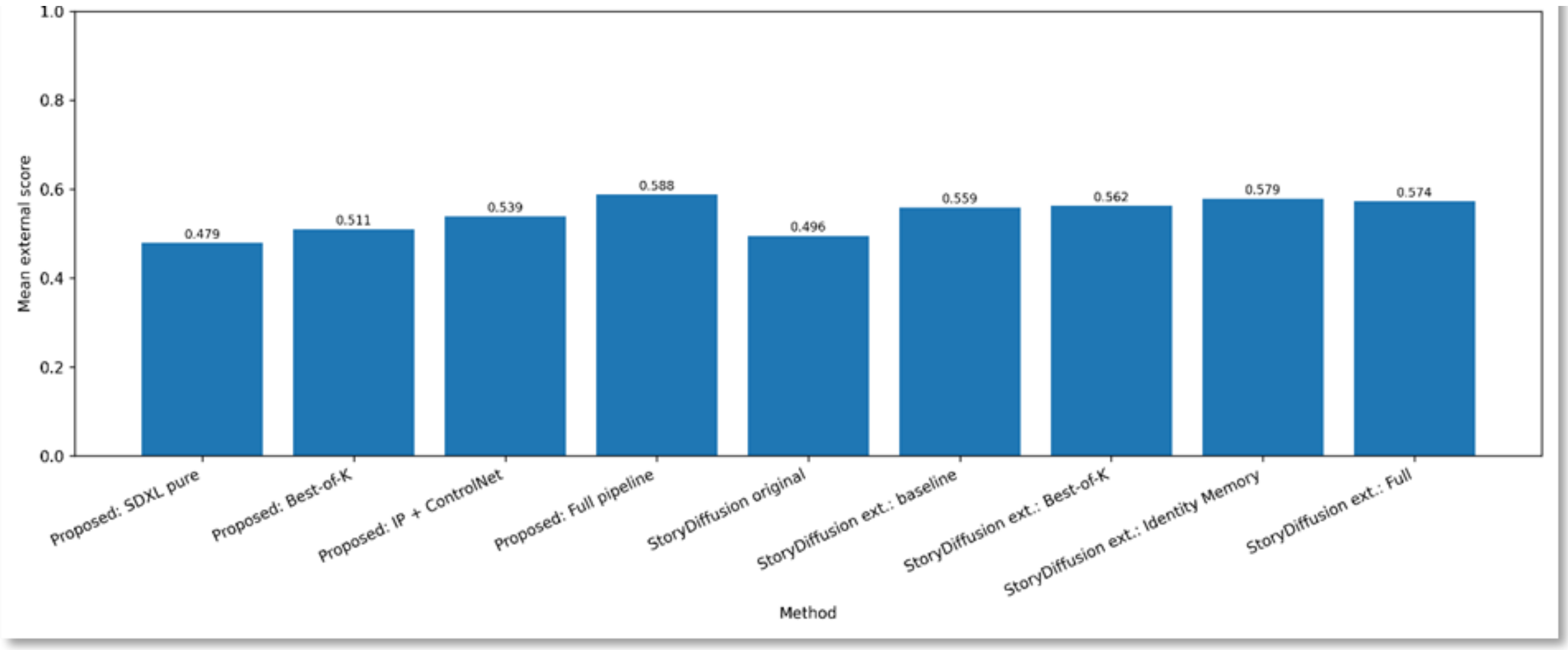

Figure 4. Overall comparison of the evaluated methods

## *D. Analysis by Difficulty and Character*

To investigate the robustness of the evaluated approaches, results were analyzed across different difficulty levels and character types.

Contrary to initial expectations, the external evaluation metrics do not decrease as story difficulty increases, as illustrated in Fig. 5. Instead, a slight increase can be observed for most methods. This behavior is likely related to the nature of CLIP and DINO embeddings, which benefit from more distinctive character representations and semantically richer prompts commonly found in difficult scenarios.

Despite this trend, the relative ranking of methods remains stable across all difficulty levels. The proposed pipeline consistently achieves the highest scores, while the original StoryDiffusion implementation remains the weakest performer.

Character-specific analysis reveals that stylized human characters such as Ana and Helena are generally easier to preserve across multiple frames. In contrast, non-human characters such as Fox Tomi and Robot Riki introduce additional challenges related to geometric structure, textures, and distinctive visual features. A detailed character-level comparison for all evaluated methods is presented in Fig. 6. These examples highlight the importance of explicit identity conditioning and memory mechanisms when generating complex cartoon characters.

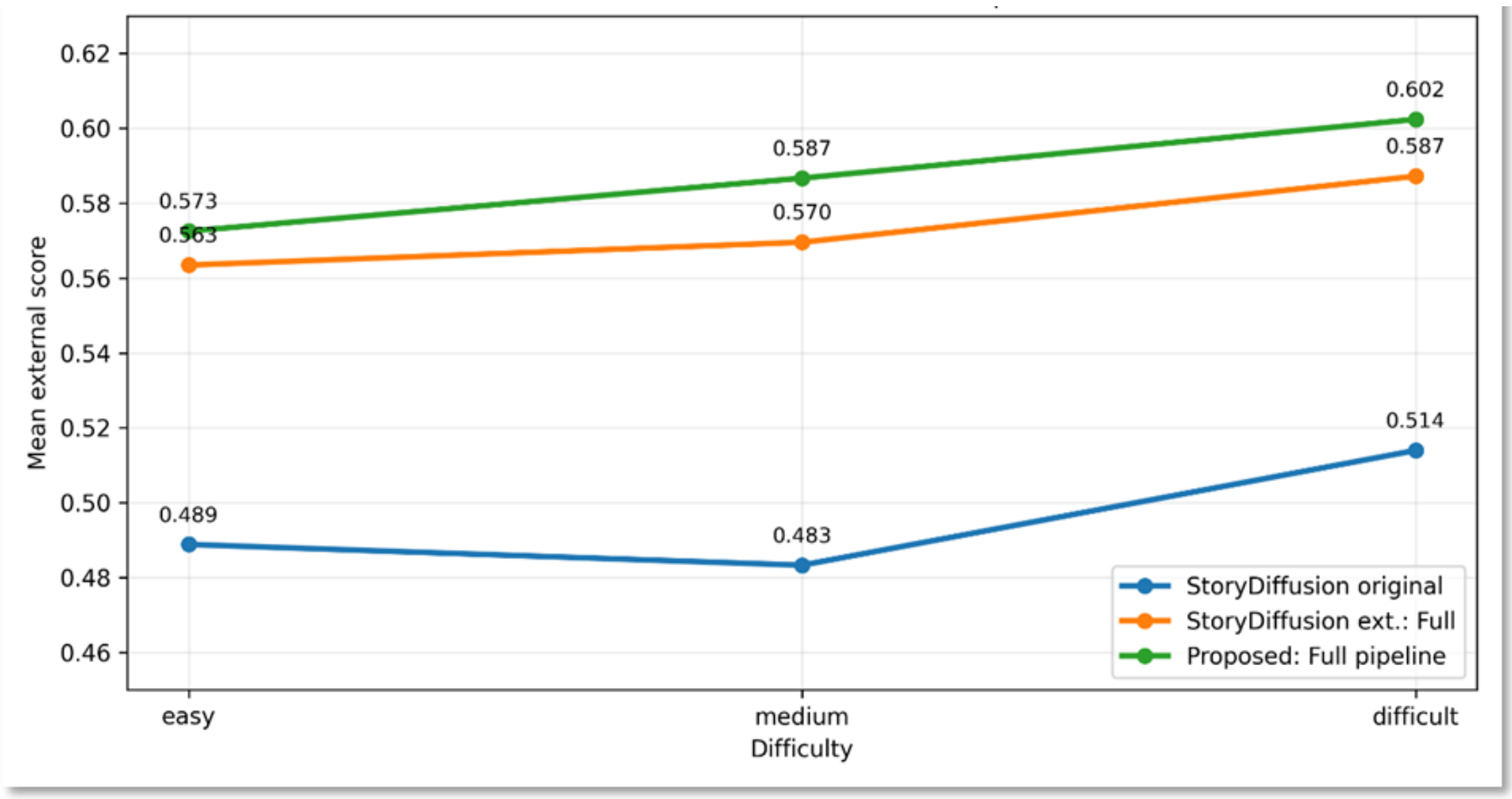

Figure 5. Performance across difficulty levels

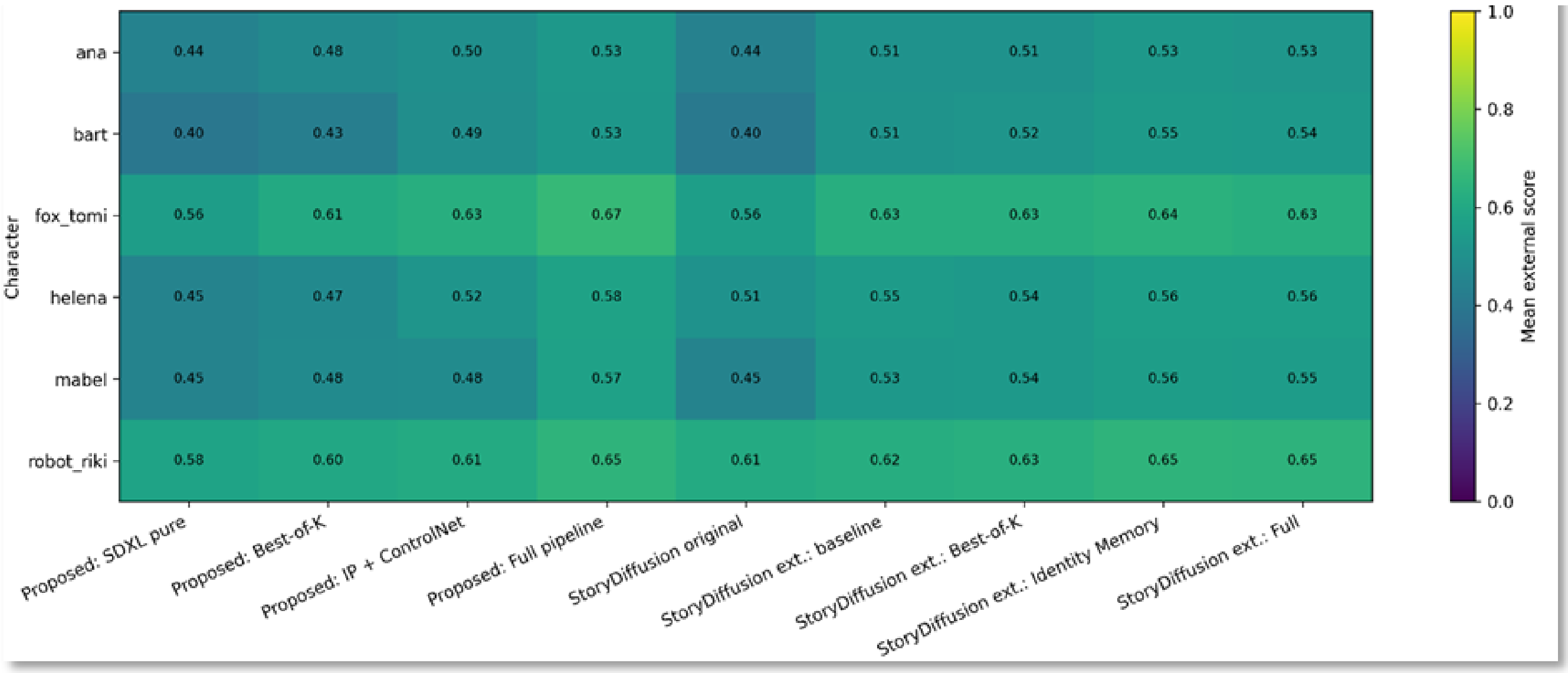

Figure 6. Character-level performance comparison

### E. *Visual comparison*

Quantitative metrics provide an objective estimate of consistency, but visual inspection remains essential for understanding the behavior of generative models.

Fig. 7 presents an easy scenario involving the character Helena. All methods preserve the general identity of the character, although noticeable differences appear in facial details and stylistic consistency. The proposed pipeline produces the most uniform sequence across the entire story.

| Reference | Character prompt | Story |
|---|---|---|
| 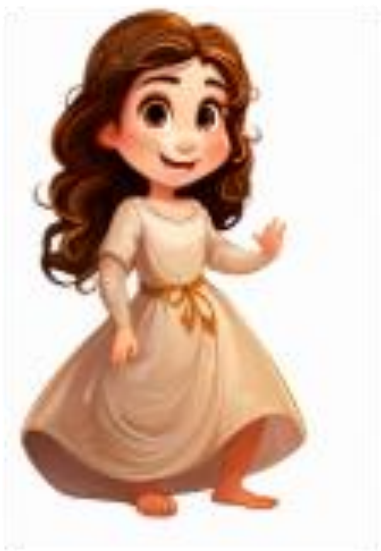 | **Identity**: young cartoon girl, long curly brown hair, large brown eyes, soft smile, **Outfit**: cream elegant dress, gold ribbon belt, light brown shoes | **Frame I** - standing at the park entrance, happy face, park gate, green trees, path, blue sky<br>**Frame II** - looking at a book on the grass, curious face, open book on grass, trees, park bench<br>**Frame III** - holding the book carefully, focused face, book in hands, park path, trees<br>**Frame IV** - walking toward a small library, determined face, small library building, sidewalk, trees<br>**Frame V** - giving the book back at the library desk, proud happy face, library desk, bookshelves, returned book |

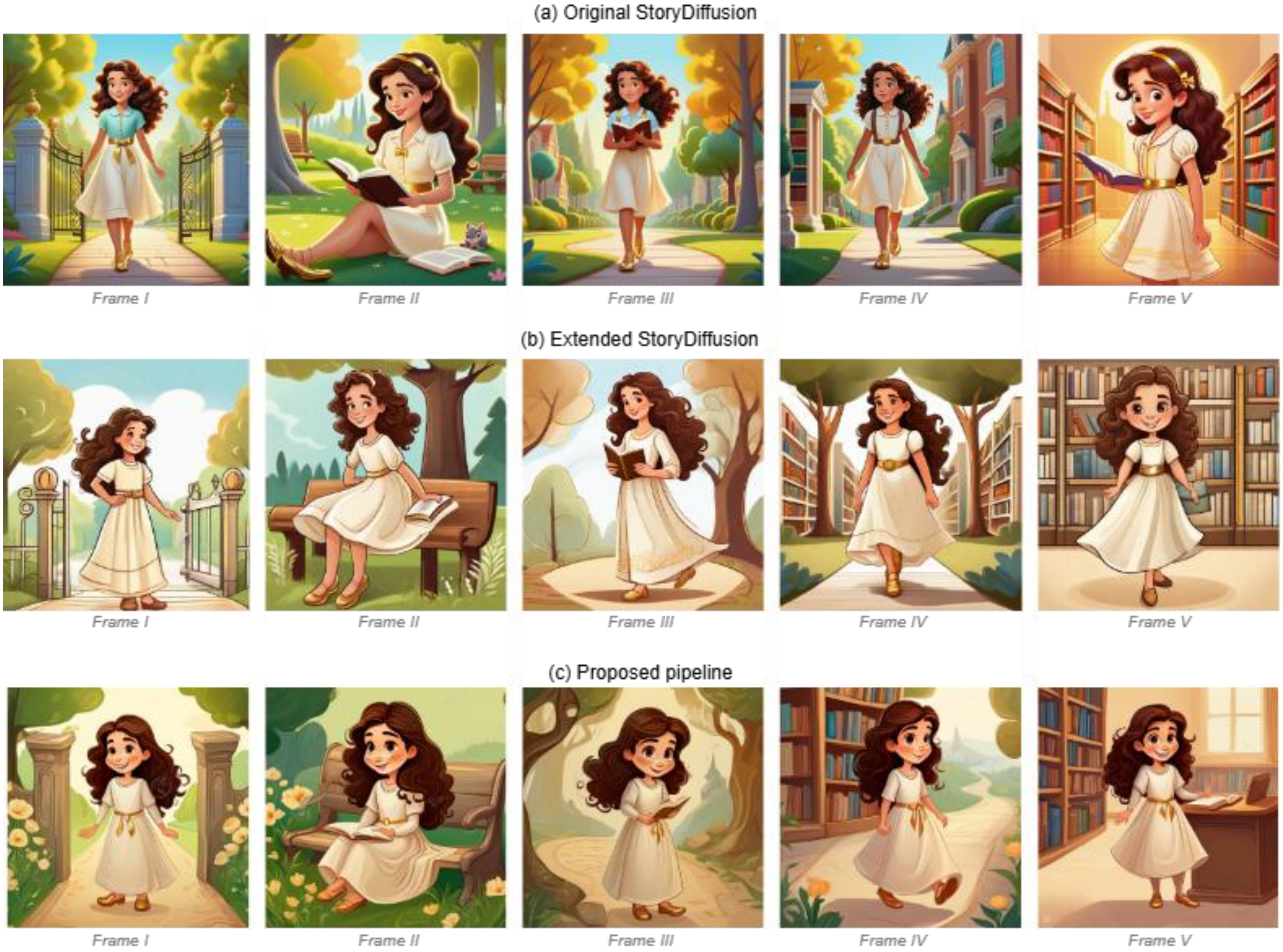

Figure 7. Visual comparison for an easy story (Helena)

Fig. 8 illustrates a medium-difficulty story involving Fox Tomi. In this example, the original StoryDiffusion baseline preserves the general appearance of the character but introduces additional elements that were not present in the original design. The extended version exhibits larger variations in proportions and stylization across frames. The proposed pipeline maintains the most coherent representation throughout the sequence.

| Reference | Character prompt | Story |
|---|---|---|
| 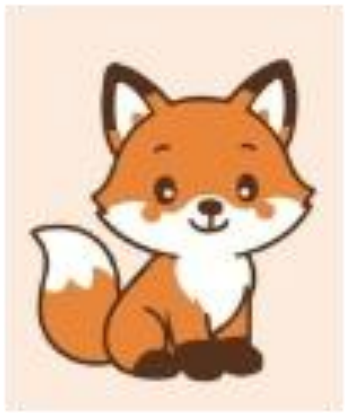 | **Identity**: friendly cartoon fox, orange fur, white muzzle, pointy ears, fluffy tail,<br>**Outfit**: — | **Frame I** - standing near a park fountain, happy face, park fountain, trees, flowers<br>**Frame II** - pointing at a wooden sign, curious face, wooden sign, park path<br>**Frame III** - picking up a dropped notebook, surprised face, notebook on grass, bench<br>**Frame IV** - running toward the library door, determined face, small library building, trees<br>**Frame V** - pushing open the library door, focused face, wooden door, books poster<br>**Frame VI** - giving the notebook at the library desk, happy face, library desk, bookshelves |

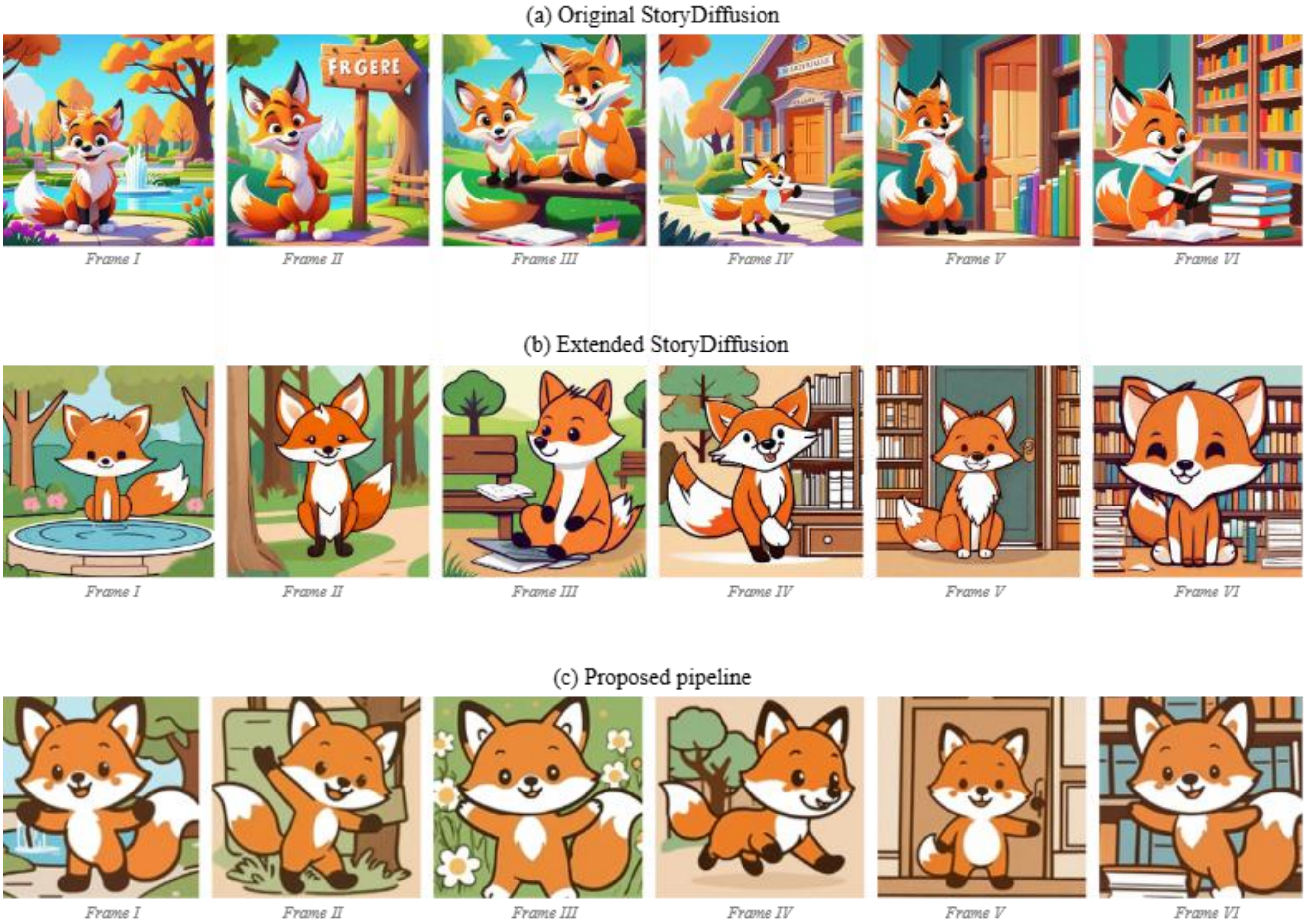

Figure 8. Visual comparison for a medium-difficulty story (Fox Tomi)

Fig. 9 presents a difficult scenario involving Bart. This example highlights the limitations of all evaluated methods. While identity drift becomes visible across all approaches, the proposed pipeline preserves the overall character structure more effectively and produces the most coherent narrative sequence.

The visual analysis confirms the conclusions obtained through quantitative evaluation and highlights the benefits of combining identity conditioning, reranking, and persistent memory mechanisms.

| Reference | Character prompt | Story |
|---|---|---|
| 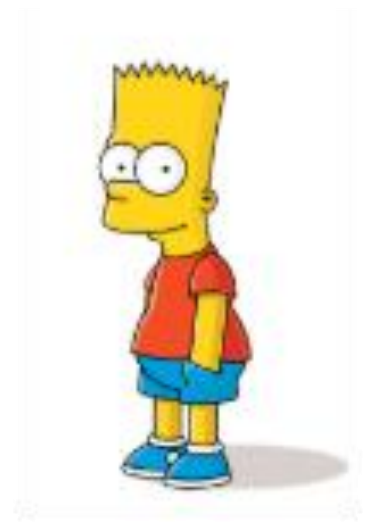 | **Identity:** cartoon boy, yellow skin, spiky yellow head, round eyes, **Outfit:** orange t-shirt, blue shorts, blue shoes, | **Frame I** - walking through a crowded city market, happy face, market stalls, lanterns, crowd<br>**Frame II** - holding a glowing crystal close to her face, surprised face, blue glowing crystal,shadows<br>**Frame III** - running in heavy rain, worried face, rain, puddles, reflections<br>**Frame IV** - peeking behind a large wooden door, nervous face, dark corridor, wooden door<br>**Frame V** - reaching toward floating magical lights, amazed face, floating lights, magical particles<br>**Frame VI** - standing in a bright library with many people, happy face, bookshelves, students, warm lighting |

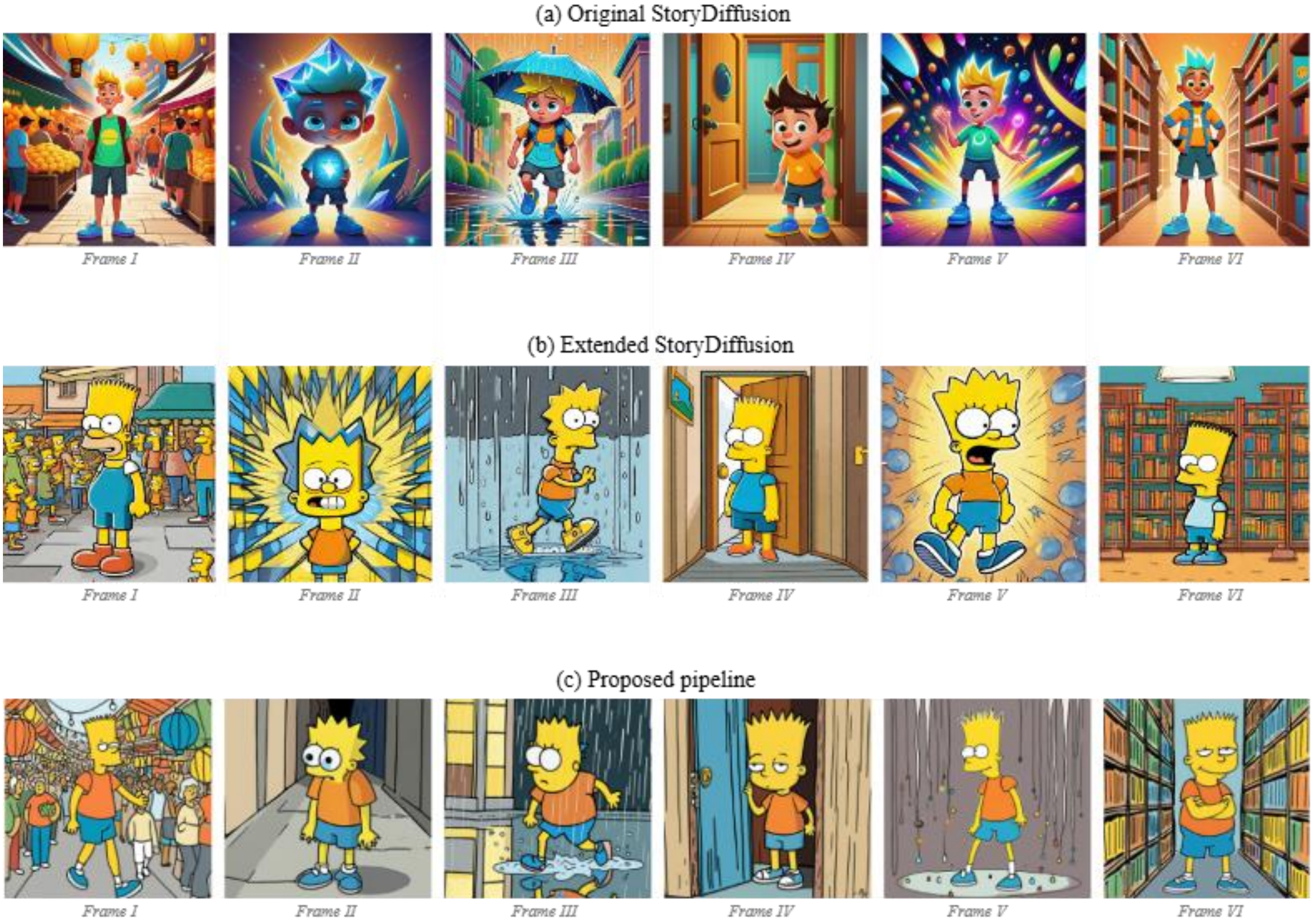

Figure 9. Visual comparison for a difficult story (Bart)

## V. Conclusion

This paper investigated the problem of character identity consistency in diffusion-based visual storytelling. While modern text-to-image models are capable of generating visually appealing images, maintaining a stable character representation across multiple narrative frames remains a challenging task. Experimental results confirmed that text-only approaches are prone to identity drift, particularly in stories involving significant changes in pose, viewpoint, illumination, or scene composition.

To address these limitations, a modular visual storytelling pipeline was proposed, combining SDXL with LoRA-based character adaptation, IP-Adapter identity conditioning, ControlNet structural guidance, Best-of-K candidate generation, automatic reranking, and a persistent Identity Memory mechanism. In addition, the StoryDiffusion framework was extended by integrating the components that provided the largest consistency improvements during preliminary experiments. A dedicated benchmark composed of multiple cartoon-style characters and stories with different difficulty levels was also developed to support systematic evaluation.

The experimental results demonstrated that both the proposed pipeline and the extended StoryDiffusion framework significantly improve character consistency compared to text-only generation. Among all evaluated methods, the complete proposed pipeline achieved the highest overall performance, producing more stable visual representations across long narrative sequences. The improvements were particularly evident in difficult scenarios, where identity drift is more likely to occur.

Despite these encouraging results, several limitations remain. Identity inconsistencies may still appear in highly complex scenes or when characters undergo substantial visual transformations. The quality of the generated results also depends on the availability of suitable reference images and carefully designed character descriptions. Furthermore, although the adopted evaluation metrics provide useful quantitative estimates, they do not always perfectly reflect human perception of visual consistency.

Future work will focus on extending the proposed methodology to stories involving multiple interacting characters, exploring more advanced temporal memory mechanisms, and evaluating the approach on larger and more diverse datasets. Another promising direction is the integration of recent multimodal foundation models capable of maintaining long-range identity information directly within the generation process. Such developments could further improve the robustness of visual storytelling systems and facilitate their application to comics, animation, interactive storytelling, and other creative domains.